\documentclass[runningheads]{llncs}
\usepackage[T1]{fontenc}
\usepackage{graphicx}
\usepackage{hyperref}
\usepackage{color}

\usepackage{amsmath} 

\usepackage{tabularx}
\usepackage{booktabs}

\newcommand\blfootnote[1]{%
  \begingroup
  \renewcommand\thefootnote{}\footnote{#1}%
  \addtocounter{footnote}{-1}%
  \endgroup
}

\begin{document}
\title{LLM-Assisted Topic Reduction for BERTopic on Social Media Data}
%
%
\author{Wannes Janssens\inst{1,2}\orcidID{0009-0009-9542-794X}
\and
Matthias Bogaert\inst{1,2}\orcidID{0000-0002-4502-0764}
\and
Dirk Van den Poel\inst{1,2}\orcidID{0000-0002-8676-8103}}
\authorrunning{W. Janssens et al.}
%
\institute{Ghent University, Research Group Data Analytics, Ghent, Belgium \and 
FlandersMake@UGent–corelab CVAMO\\
\email{wanjanss.janssens@UGent.be}
}
\maketitle              
\begin{abstract}
The BERTopic framework leverages transformer embeddings and hierarchical clustering to extract latent topics from unstructured text corpora. While effective, it often struggles with social media data, which tends to be noisy and sparse, resulting in an excessive number of overlapping topics. Recent work explored the use of large language models for end-to-end topic modelling. However, these approaches typically require significant computational overhead, limiting their scalability in big data contexts. In this work, we propose a framework that combines BERTopic for topic generation with large language models for topic reduction. The method first generates an initial set of topics and constructs a representation for each. These representations are then provided as input to the language model, which iteratively identifies and merges semantically similar topics. We evaluate the approach across three Twitter/X datasets and four different language models. Our method outperforms the baseline approach in enhancing topic diversity and, in many cases, coherence, with some sensitivity to dataset characteristics and initial parameter selection.\blfootnote{Presented at the NFMCP, ECML PKDD 2025 Workshop}

\end{abstract}

\section{Introduction}

A popular goal in social media data analysis is the discovery of underlying trends or topics, either to gain insights into online discourse and customer experiences \cite{egger2022topic}, or for use in downstream applications \cite{janssens2021evaluating,yu2019hierarchical,crijns2023topic}. Topic modelling is a widely used technique for uncovering hidden structure in unstructured text \cite{blei2009topic}. However, analysing social media data presents unique challenges. These data sources typically arrive in large volumes and consist of short, noisy, and sparse texts, making traditional topic modelling techniques less effective \cite{laureate2023systematic}. Methods such as Latent Dirichlet Allocation (LDA) and Non-negative Matrix Factorization (NMF), which rely on bag-of-words representations, often fail to capture meaningful patterns due to the limited word co-occurrence in short texts \cite{qiang2020short}.

BERTopic \cite{grootendorst2022bertopic} offers a more advanced alternative. It uses Sentence-BERT (SBERT) \cite{reimers2019sentence} embeddings to capture semantic information of documents and applies HDBSCAN \cite{malzer2020hybrid}, a hierarchical and density-based clustering algorithm, to group documents into topics. While BERTopic outperforms traditional methods in many aspects, it often produces an excessive number of topics when applied to social media data. These topics are not only numerous but also frequently overlapping. Recently, large language models (LLMs) have been investigated as an alternative for end-to-end topic modelling, leveraging their strong semantic understanding to extract topics from various sources \cite{stammbach2023re,li2023can,wang2023prompting,pham2023topicgpt}. However, these approaches typically process each input document sequentially, resulting in high computational costs that make them impractical for large-scale social media datasets.

Our proposed method leverages the best of both worlds by (1) first generating an initial set of topics with BERTopic, and (2) using LLMs as assistant to effectively reduce the number of topics. For topic reduction, each topic is represented either by its top 10 keywords, extracted using c-TF-IDF, or by a label generated by an LLM. These topic representations are provided to an LLM, which is prompted to iteratively identify and merge overlapping topics until a desired number of topics is reached. We evaluate this approach by comparing the coherence and diversity of the reduced topics across three social media datasets (Trump Tweets, \#Covid19 Tweets and Cyberbullying Tweets) using four different LLMs: GPT-4o-mini, Llama3-8b, Gemma3-12b and Qwen3-30b-a3b. As a baseline, we compare it to a built-in topic reduction method for BERTopic, which reduces the number of topics by increasing the \textit{minimum cluster size} parameter in HDBSCAN. Our results show that the LLM-assisted topic reduction process yields more diverse and in many cases more coherent topics compared to the baseline while achieving a better balance between quality and computational efficiency compared to end-to-end LLM approaches.  

\section{Related work}

\subsection{BERTopic and Topic Reduction}

BERTopic is a topic modelling framework composed of four modular components. It starts with Sentence-BERT (SBERT) to generate contextual embeddings for input documents. These high-dimensional embeddings are then reduced using UMAP \cite{mcinnes2018umap}, a non-linear dimensionality reduction technique. The reduced embeddings are clustered using HDBSCAN \cite{malzer2020hybrid}, a hierarchical density-based clustering algorithm. Finally, class-based TF-IDF (c-TF-IDF) combined with CountVectorizer is employed to extract representative keywords for each topic. BERTopic has gained traction due to its flexible, modular design, which allows users to substitute different components at each step. Moreover, it has demonstrated superior performance compared to traditional topic modelling methods such as LDA and NMF in various studies \cite{ogunleye2023comparison,abuzayed2021bert,de2022experiments}, especially in terms of topic coherence and clustering quality \cite{gan2023experimental,turan2023comparison}.

Among its components, HDBSCAN plays a particularly critical role. It generates clusters without requiring the number of topics to be specified in advance. The algorithm constructs a hierarchy of clusters and prunes it to produce the final set. However, when applied to social media data, which is often noisy, sparse, and short in nature, HDBSCAN tends to produce an excessive number of topics, many of which are semantically overlapping. This motivates the need for post hoc topic reduction to improve interpretability and manageability. In practice, many studies using BERTopic, whether to analyse latent trends or compare topic modelling techniques, apply some form of topic reduction. Still, few describe this process in detail. Common approaches include using BERTopic’s built-in \textit{nr\_topics} parameter to limit the number of topics or resorting to manual pruning or ad hoc merging strategies \cite{egger2022topic,baird2022consumer}. This indicates a clear lack of methodological transparency and standardization, which underscores the need for reproducible topic reduction strategies, particularly for challenging datasets like social media.


\subsection{LLMs for Topic Modelling}

Recent advances in LLMs have sparked interest in their application to topic modelling. These applications generally fall into two categories: end-to-end topic modelling using LLMs, and the use of LLMs to assist or enhance specific steps within the topic modelling process.

PromptTopic \cite{wang2023prompting} and TopicGPT \cite{pham2023topicgpt} are two approaches that rely on LLMs for end-to-end topic modelling. PromptTopic generates sentence-level topics using prompt-based methods, merges redundant topics, and refines them by selecting representative keywords. TopicGPT follows a similar strategy, first generating topics from a document sample and then assigning and refining them across the corpus using embedding-based merging. Doi et al. \cite{doi2024topic} propose another end to end topic modelling framework in which large language models are prompted either in parallel or sequentially to extract and merge topics.

Although these methods demonstrate that large language models can yield coherent and high level topics, they also face notable computational limitations. In particular, they require processing each document or even parts of documents individually because of input length constraints, which leads to prohibitive costs when applied to large scale datasets such as social media corpora. Furthermore, their global corpus wide approach can reduce granularity and overlook nuanced or niche themes that are often crucial in short text analysis. Finally, their reported results are not consistently superior to alternative topic modelling methods, and their comparisons do not account for the sensitivity of BERTopic to its parameters. In our work we explicitly addressed this by averaging results across multiple HDBSCAN configurations.

In contrast to end-to-end approaches, other research has focused on LLM-assisted topic modelling, where LLMs are used to enhance specific components of traditional pipelines. For instance, Stammbach et al. \cite{stammbach2023re} integrated LLMs like ChatGPT and FLAN-T5 for topic evaluation, using intrusion detection and coherence rating tasks. Their results show that LLMs can reliably assess topic quality and even help determine the optimal number of topics in models like LDA. Similarly, Li et al. \cite{li2023can} used ChatGPT to generate topic labels from LDA outputs. Their study found that LLM-generated labels were frequently preferred over human-authored ones, highlighting the potential of LLMs to improve topic interpretability. Koloski et al. \cite{koloski2024aham} also employ LLMs to create topic labels that are more domain specific, showing that prompts guided by domain experts can yield labels that better reflect the context of the data. In addition, they introduce a new evaluation metric and explore a domain adaptations for BERTopic by fine tuning sentence transformer embeddings to better match a given domain. However, they do not tune BERTopic parameters or apply any post hoc topic reduction. These studies reflect a growing interest in LLMs for tasks such as evaluation and representation refinement.

To our knowledge, no prior work has explored the use of LLMs for topic reduction by merging semantically overlapping or redundant topics in transformer-based frameworks like BERTopic. Our work addresses that gap by combining BERTopic’s ability to generate fine-grained topics with the semantic reasoning capabilities of LLMs to reduce topic redundancy, resulting in a hybrid approach with significantly lower computational cost than end-to-end LLM-based alternatives.

\section{Methodology} \label{proposed_framework}

\begin{figure}
    \centering
    \includegraphics[width=1\linewidth]{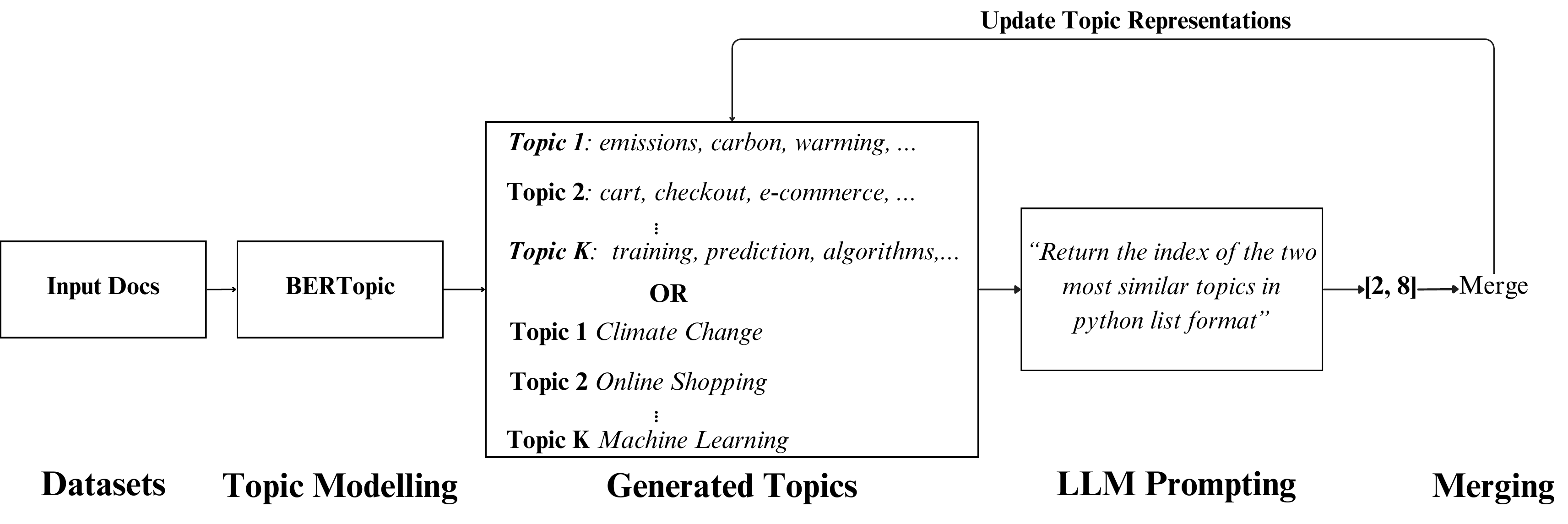}
    \caption{LLM-Assisted Topic Reduction: Topics are generated by BERTopic and reduced iteratively by prompting an LLM. After each merge topic representations are updated for all topics}
    \label{fig:LLM_assistence}
\end{figure}

We propose a simple yet effective topic modelling methodology that combines the topic generation strengths of BERTopic with the semantic reasoning and interpretative capabilities of LLMs for topic reduction as illustrated in Figure \ref{fig:LLM_assistence}.

In the first step, we apply the default configuration of the BERTopic framework to generate an initial set of topics from a given input corpus. To explore topic granularity, we vary key parameters of HDBSCAN, such as the \textit{minimum cluster size} and \textit{cluster selection method}, which influence the number and density of resulting clusters. Once the topics are generated, we construct topic representations using one of two approaches:
(1) the default method of extracting the top 10 keywords per topic using c-TF-IDF, or
(2) using an LLM to generate a concise label for each topic based on the documents and keywords associated with it. These topic representations are then fed into an LLM, which is prompted to identify the two most semantically overlapping topics from the set. In an iterative reduction loop, the selected topics are merged, a new topic representation is generated, and the process is repeated, until a user-defined target number of topics is reached. While we considered single-shot clustering, we believe that iterative agglomerative merging offers finer control over the number of topics and helps prevent overwhelming the LLM with very large input lists, nonetheless, single-shot clustering remains an interesting avenue for future research. This LLM-assisted merging introduces semantic reasoning into the reduction process, allowing the model to consider meaning beyond surface-level keyword overlap.

\begin{figure}
    \centering
    \includegraphics[width=0.87\linewidth]{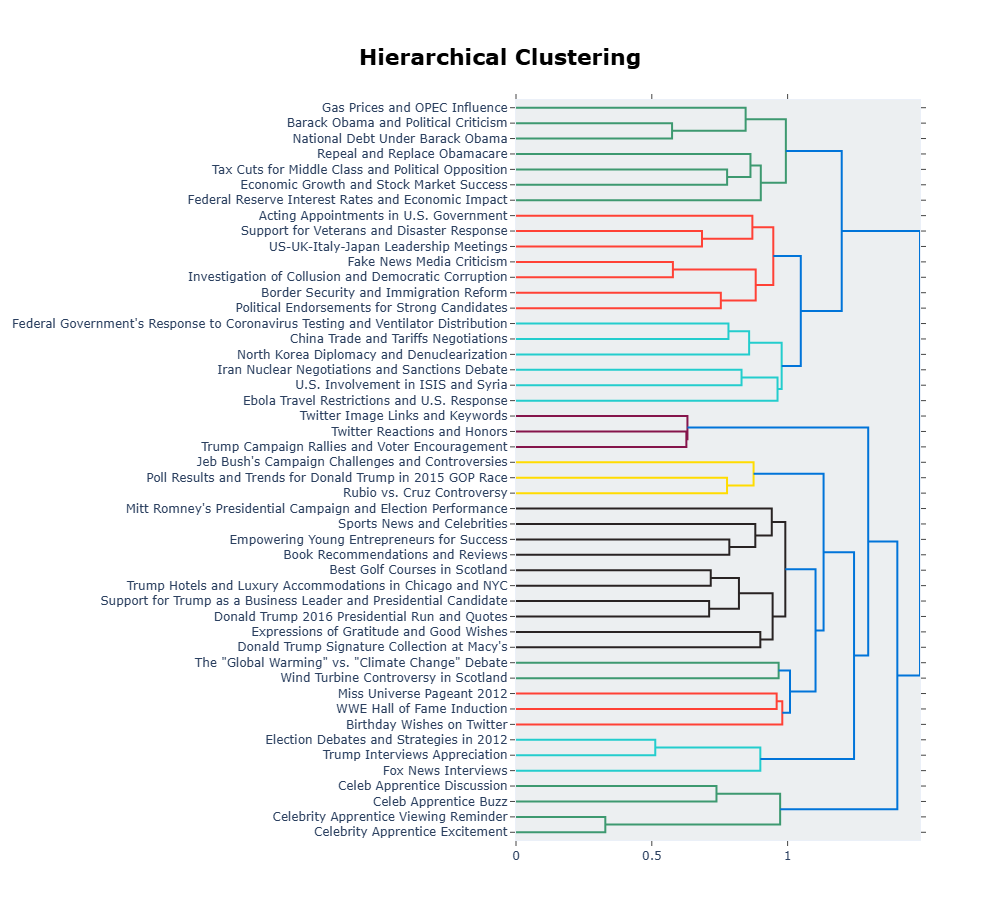}
    \caption{Hierarchical representation of (48) topics generated by BERTopic on the Trump Tweets dataset BEFORE applying our proposed reduction method. Topic labels were generated by an LLM to enhance interpretability}
    \label{fig:before}
\end{figure}

\begin{figure}
    \centering
    \includegraphics[width=0.87\linewidth]{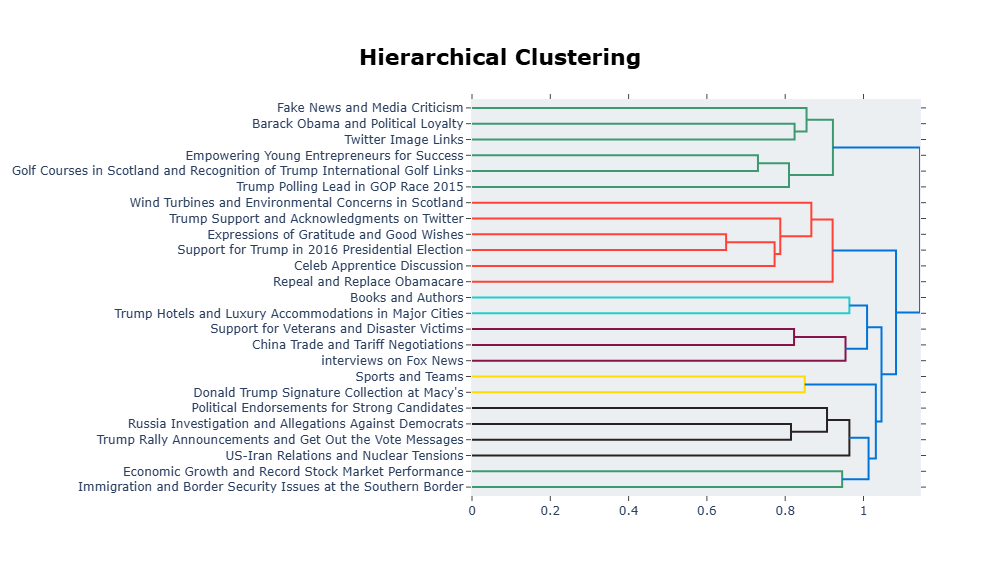}
    \caption{Hierarchical representation of (25) topics AFTER applying our proposed reduction method on the Trump Tweets dataset. Topic labels were generated by an LLM to enhance interpretability}
    \label{fig:after}
\end{figure}

\begin{figure}
    \centering
    \includegraphics[width=0.8\linewidth]{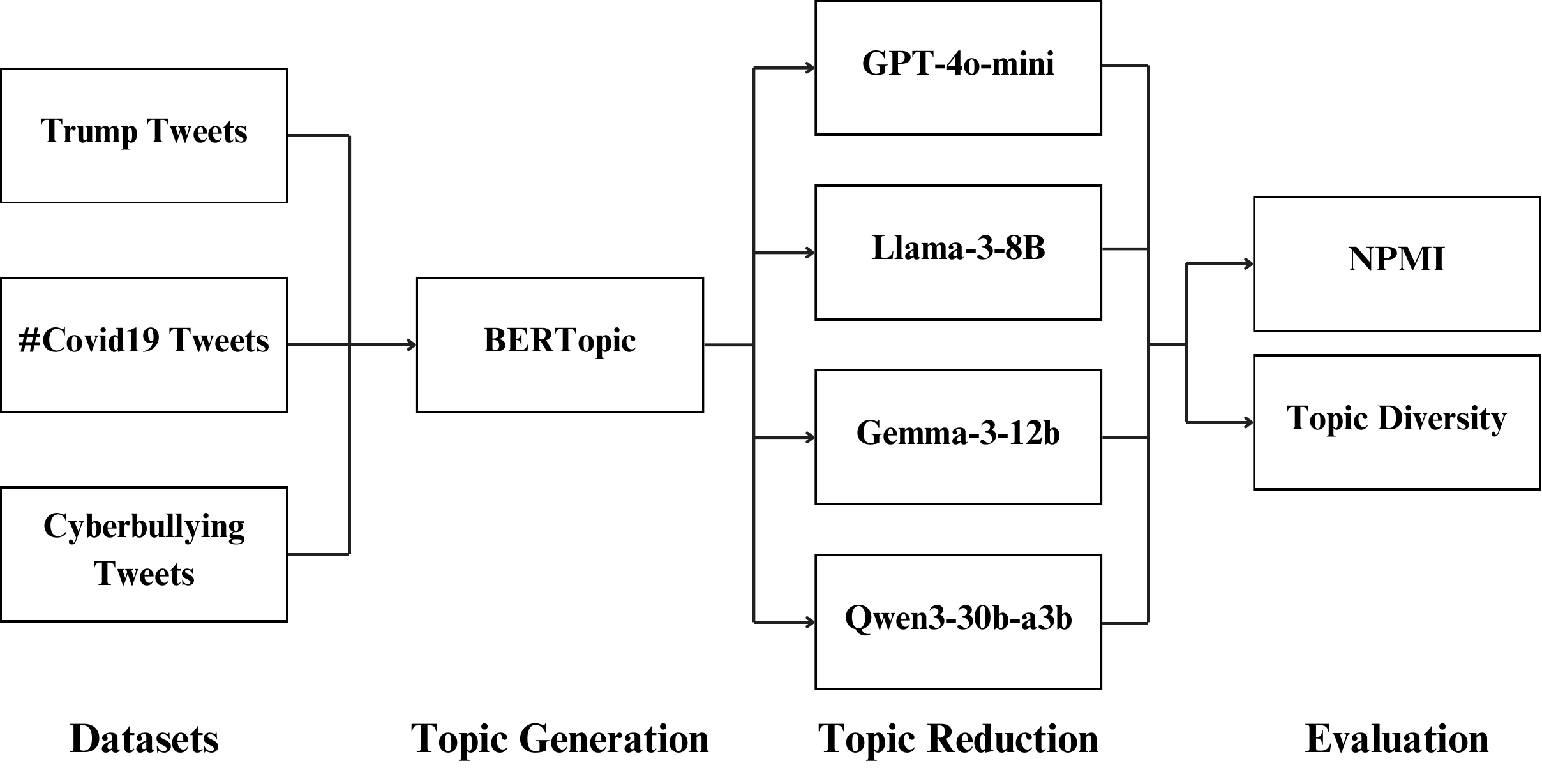}
    \caption{Experimental setup, including datasets, LLMs and evaluation metrics}
    \label{fig:exp_setup}
\end{figure}

An illustration for the Trump Tweets dataset is provided: Figure \ref{fig:before} shows the topics generated by the BERTopic framework before the LLM assisted reduction, and Figure \ref{fig:after} shows the reduced topic set. To validate our methodological framework, a full overview of our experimental setup is given in Figure \ref{fig:exp_setup} and implementation details are provided in following sections. 

\subsection{Data}
The topic reduction method is evaluated using three open-source datasets collected from Kaggle: Trump Tweets\footnote{\url{https://www.kaggle.com/datasets/austinreese/trump-tweets}}, \#Covid19 Tweets\footnote{\url{https://www.kaggle.com/datasets/gpreda/covid19-tweets}} and Cyberbullying Tweets\footnote{\url{https://www.kaggle.com/datasets/andrewmvd/cyberbullying-classification}}. The Trump Tweets dataset contains 43,352 posts collected from the @realDonaldTrump Twitter account, covering the period from May 2009 to June 2020. The \#Covid19 Tweets dataset comprises 178,683 tweets posted between June and August 2020, each tagged with the hashtag \textit{Covid19}. Finally, the Cyberbullying Tweets dataset contains about 39,747 posts related to various sorts of cyberbullying on X. Table \ref{tab:datasets} presents a summary of the key features of these datasets. To ensure fair comparisons and retain maximum contextual information for the SBERT embedding module used in the BERTopic framework, the datasets are left unprocessed.

\begin{table}[h]
    \caption{Dataset size and average document length}
    \centering
    \begin{tabular}{lccc}
        \toprule
        \textbf{Dataset} & \textbf{Size (\#docs)} & \textbf{Avg \#Words} & \textbf{Avg \#Characters}\\
        \midrule
        Trump Tweets & 43,352  & 20.76 & 131.53\\
        \#Covid19 Tweets & 179,108 & 17.61 & 130.52 \\
        Cyberbullying Tweets & 39,747 & 25.27 & 146.88 \\
        \bottomrule
    \end{tabular}
    \label{tab:datasets}
\end{table}

\subsection{BERTopic Configuration}
\begin{figure}
    \centering
    \includegraphics[width=1\linewidth]{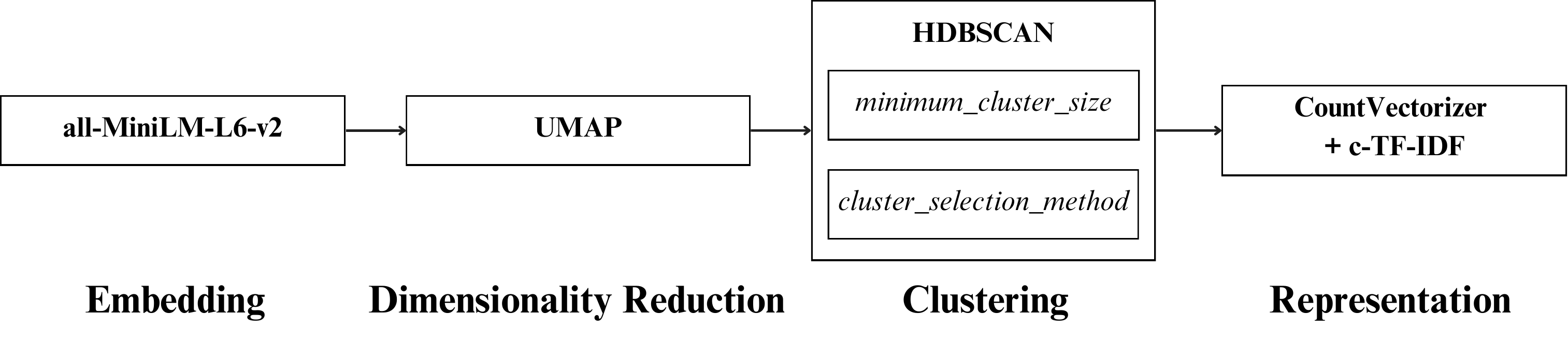}
    \caption{The BERTopic framework with the chosen modules for each step. Default parameters were used for each module, except for HDBSCAN}
    \label{fig:BERTopic_setup}
\end{figure}

Each dataset is processed by the BERTopic framework, which is composed of four consecutive modules: embedding, dimensionality reduction, clustering, and topic representation. Although the framework supports customization for each component, we employed the default configuration due to its proven effectiveness and common adoption. This default setup utilizes the pre-trained \textit{"all-MiniLM-L6-v2"} model from SBERT for generating embeddings, UMAP for reducing dimensionality, HDBSCAN for clustering, and a representation module that integrates CountVectorizer with c-TF-IDF, as depicted in Figure \ref{fig:BERTopic_setup}.

Instead of relying on the default HDBSCAN configuration, we adjust the \textit{minimum cluster size} and \textit{cluster selection method} parameters, as they have a notable impact on clustering outcomes. Specifically, \textit{minimum cluster size} controls the smallest group size considered for forming a cluster during the tree-building process, while \textit{cluster selection method} determines whether the final clusters are selected from the tree’s leaf nodes or based on overall cluster stability. 

\subsection{LLMs}
After generating an initial set of topics with BERTopic, we apply LLM-assisted topic reduction. While numerous LLMs with varying characteristics are available, we selected the following subset: GPT-4o-mini, Llama-3-8B, Gemma-3-12B, and Qwen-3-30B-A3B. This selection provides a mix of open/closed source, model sizes, and architectural diversity. GPT-4o-mini is a compact proprietary model optimized for low latency. Llama-3-8B and Gemma-3-12B are open-weight decoder-based models released by Meta and Google. Qwen-3-30B-A3B is a large-scale sparse mixture-of-experts model from Alibaba, featuring 64 experts with 8 active per token. A summary of the chosen models is presented in Table \ref{tab:llms}. The GPT model was accessed via the OpenAI API, while the LLaMA, Gemma, and Qwen models were run locally using Ollama.

\begin{table}
\centering
\caption{Comparison of LLMs used in this study}
\label{tab:llms}
\begin{tabular}{@{} l c p{0.25\textwidth} p{0.4\textwidth} @{}}
\toprule
\textbf{Model} & \textbf{Size} & \textbf{Architecture} & \textbf{Remarks} \\
\midrule
GPT-4o-mini    & Unknown & Proprietary (OpenAI) & Compact GPT-4o variant, optimized for latency \\
Llama-3-8B     & 8B      & Open-weight decoder (Meta) & Optimized for instruction following \\
Gemma-3-12B    & 12B     & Open-weight decoder (Google) & Based on Gemini architecture, optimized for efficiency \\
Qwen-3-30B-A3B & 30B     & Open-weight, Sparse MoE (Alibaba) & Sparse mixture of experts model with 64 experts, 8 active per token \\
\bottomrule
\end{tabular}
\end{table}

\subsection{Evaluation}
To ensure a fair and objective evaluation of the topic reduction methods, we extract the top 10 keywords of each topic using c-TF-IDF and employ automated metrics. Topic coherence is measured using Normalized Pointwise Mutual Information (NPMI), which ranges from ‑1 to 1, with 1 indicating perfect co‑occurrence and ‑1 indicating complete divergence. NPMI has been shown to exhibit the highest correlation with human judgment \cite{roder2015exploring} compared to other coherence measures. To assess topic diversity, we compute the ratio of unique words in the top keywords across all topics to the total number of topic words. These metrics provide both local and global assessments of topic quality. 


For each dataset, the coherence and diversity scores are averaged across four runs by combining two different \textit{minimum cluster size} values with the two available \textit{cluster selection methods}. Specifically, we use minimum cluster sizes of 100 and 150 for the Trump Tweets dataset, 200 and 300 for the \#Covid19 Tweets dataset, and 40 and 50 for the Cyberbullying Tweets dataset. These settings result in varying numbers of initial topics across datasets. 

\section{Results and Discussion}
For each dataset we run four different BERTopic configurations with varying \textit{minimum cluster size} and \textit{cluster selection method} parameters. Next, the LLM-assisted topic reduction method decreases the number of topics to two pre-defined amounts for each dataset (50 and 25 for Trump tweets, 100 and 50 for the \#Covid19 tweets and 30 and 15 for the cyberbullying tweets). Coherence and topic diversity of the reduced topics for each LLM are compared with increasing the \textit{minimum cluster size} in BERTopic, a baseline method which is commonly used for topic reduction in BERTopic. 

It is worth noting that Llama-3 often failed to return responses, particularly when using LLM-generated labels due to the presence of sensitive content (e.g., racism, antisemitism, sexism, sexual violence, and homophobia). Especially, the cyberbullying dataset contained highly sensitive content. As many LLMs are trained to avoid responding to or reproducing harmful content, this can limit their effectiveness when tasked with analysing sensitive domains.

\subsection{Trump Tweets}
\begin{figure}
    \centering
    \includegraphics[width=1\linewidth]{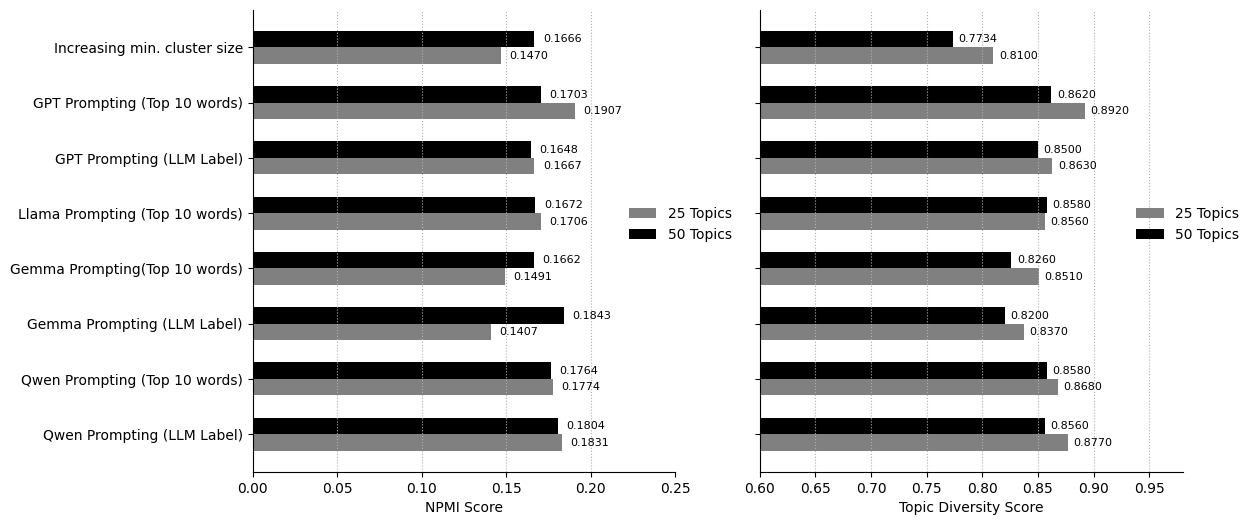}
    \caption{Results Trump Tweets}
    \label{fig:results_trump}
\end{figure}

The results for the Trump Tweets dataset are given in Figure \ref{fig:results_trump}. When reducing to 50 topics, both Gemma and Qwen (especially using LLM labels) achieve the highest topic coherence. GPT performs slightly better when using the top-10 words prompt format, while the Gemma top-10 words and the GPT LLM-label setup do not show significant improvements compared to the baseline of increasing the \textit{minimum cluster size}. Upon reducing to 25 topics, Qwen maintains strong coherence, while performance slightly declines for most other models. Interestingly, GPT with top-10 words shows a small increase in coherence at 25 topics, whereas Gemma (LLM label) experiences a notable drop, underperforming compared to the baseline.

In terms of topic diversity, all LLM-based methods outperform the baseline. Diversity improves consistently when reducing further to 25 topics. Qwen and GPT stand out as the top performers in this dataset, achieving a strong balance between coherence and diversity.

\subsection{\#Covid19 Tweets}
\begin{figure}
    \centering
    \includegraphics[width=1\linewidth]{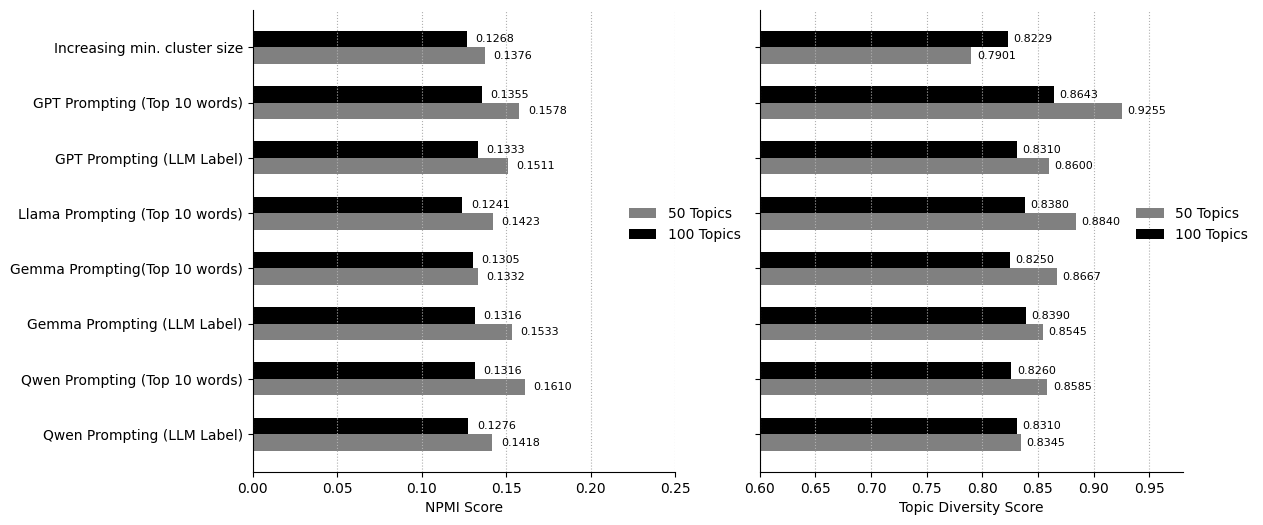}
    \caption{Results \#Covid19 Tweets}
    \label{fig:results_covid}
\end{figure}

Figure \ref{fig:results_covid} shows the results on the \#Covid19 Tweets dataset. At 100 topics, all LLM methods except Llama slightly outperform the baseline in terms of coherence. However, at 50 topics, coherence improves substantially for all models, especially for Qwen (top-10 words), GPT (LLM label), and Gemma (LLM label). In this setting, all LLM methods clearly outperform the baseline.

Diversity-wise, all models show higher topic diversity than the baseline when reducing to 100 topics. When further reduced to 50 topics, diversity increases significantly across the board, with GPT (top-10 words) emerging as the most effective method in this context.

\subsection{Cyberbullying Tweets}
The results for the cyberbullying dataset reveal a distinct pattern compared to the Trump and \#Covid19 datasets. In terms of topic coherence, all LLM-based reduction methods underperform relative to the baseline, both when reducing to 30 and 15 topics. Among the LLMs, GPT (LLM label) performs best at 30 topics, while Gemma (top-10 words) shows the strongest coherence at 15 topics. This suggests that LLMs struggle to maintain semantic consistency in this domain, likely due to the nature of the dataset.

\begin{figure}
    \centering
    \includegraphics[width=1\linewidth]{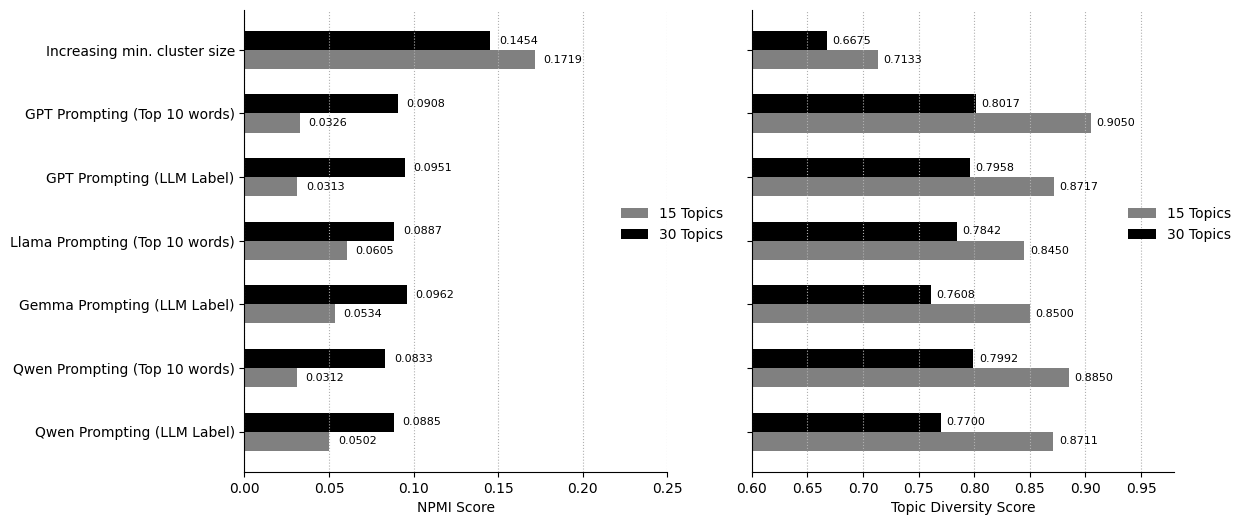}
    \caption{Results Cyberbullying Tweets}
    \label{fig:results_cyber}
\end{figure}

In contrast, the results for topic diversity present an entirely different picture. Here, all LLM-based methods significantly outperform the baseline approach, at both reduction levels. GPT (top-10 words) achieves the highest diversity scores overall, followed closely by Qwen and Gemma. This indicates that while LLMs may struggle with semantic precision in sensitive or noisy domains, they still excel at capturing a broad range of distinct themes.

\section{Conclusion and Further Research}
Our results across three diverse datasets demonstrate that LLM-assisted topic reduction can offer clear advantages over traditional BERTopic reduction strategies, particularly in enhancing topic diversity and, in many cases, topic coherence. On the Trump and \#Covid19 datasets, models such as GPT-4o-mini, Qwen, and Gemma produced strong results, especially when using prompt formats based on top-10 keywords. This approach eliminates the need to generate LLM topic labels at each iteration, thereby reducing computational overhead. However, on the Cyberbullying dataset, while LLMs consistently improved topic diversity, they underperformed in terms of coherence compared to the baseline. This suggests that the effectiveness of LLM-based topic reduction may be dataset-dependent, potentially influenced by the nature or complexity of the underlying content. Additionally, models such as Llama-3 frequently refused to respond when prompted with data from the Cyberbullying dataset, likely due to the sensitive nature of the content. This behaviour points to possible limitations in handling harmful or sensitive topics, possibly resulting from training data gaps or alignment constraints that restrict model outputs in such contexts.

Future work could explore more advanced prompting strategies to enhance the identification and merging of overlapping topics, such as dynamic or context-aware prompting. Moreover, a comparison with other LLM‑based topic modelling approaches  and additional topic reduction techniques beyond simply increasing the minimum cluster size would provide a more comprehensive evaluation. Additionally, a more in-depth investigation into the impact of initial parameter settings within the BERTopic framework could provide further insights into optimal topic generation. Finally, incorporating human evaluation may offer a more nuanced assessment of topic quality compared to reliance on automated metrics alone.

\begin{credits}

\subsubsection{\discintname}
The authors have no competing interests to declare that are
relevant to the content of this article.
\end{credits}

%
%
%
\bibliographystyle{splncs04}
\bibliography{references}

\end{document}